%% file: template_CGIconf.tex
\RequirePackage{fix-cm}
\documentclass[twocolumn]{svjour3}          
\smartqed  
\usepackage{url}
\usepackage{graphicx}
\usepackage{amsmath}
\usepackage{amsfonts}
\usepackage{xcolor}
\usepackage{tabularx}
\usepackage{multicol}
\usepackage{multirow}
\usepackage{comment}
\usepackage{float}
\usepackage[caption=false,font=normalsize,labelfont=rm,textfont=rm]{subfig}
\usepackage{diagbox}
\journalname{CGI2025} 
\begin{document}

\title{IG-Diff: Complex Night Scene Restoration with Illumination-Guided Diffusion Model}
\author{Yifan Chen \and Fei Yin \and Chunle Guo \and Chongyi Li \and Yujiu Yang}
\institute{Yifan Chen \and Fei Yin \and Yujiu Yang \at Tsinghua Univerisity
\and Chunle Guo \and Chongyi Li \at 
NanKai University}
\date{ }

\maketitle
\sloppy 
\begin{abstract}
\input{main/0_abstract}
\end{abstract}

\input{main/1_introduction}
\input{main/2_relatedworks}
\input{main/3_dataset}

\input{main/3_proposedmethod}
\input{main/4_experiment}
\input{main/5_conclusion}

\bibliographystyle{spmpsci}
\bibliography{reference}



\end{document}

%% file: main/0_abstract.tex
In nighttime circumstances, it is challenging for individuals and machines to perceive their surroundings.
While prevailing image restoration methods adeptly handle singular forms of degradation, they falter when confronted with intricate nocturnal scenes—such as the concurrent presence of weather and low-light conditions.
Compounding this challenge, the lack of paired data that encapsulates the coexistence of low-light situations and other forms of degradation hinders the development of a comprehensive end-to-end solution.
In this work, we contribute complex nighttime scene datasets that simulate both illumination degradation and other forms of deterioration. 
To address the complexity of night degradation, 
%
we propose an integration of an illumination-guided module embedded in the diffusion model to guide the illumination restoration process.
Our model can preserve texture fidelity while contending with the adversities posed by various degradation in low-light scenarios.
%

%% file: main/1_introduction.tex
\section{Introduction}
\input{tex/image_tex/weather/teaser_weather}

Image restoration amidst challenging conditions, such as adverse weather like rain or fog, and complicated degradation like blur or noise, poses a practical issue within the realms of surveillance and automotive sectors.
Real-world scenarios often couple these conditions with the constrictions of low-light environments caused by insufficient illumination or the constraints of limited exposure time. 
%
When these unfavorable atmospheric circumstances intermingle with the constraints of lowlight, the resultant effect constitutes a formidable obstacle to human perception and downstream tasks~\cite{li2021lowlightsurvey}.
To address this issue, complex night scene restoration techniques have been proposed, which aim to manipulate color, contrast, and obstruction to recover the visual quality of these images. 

Previous methods have typically addressed the tasks of low-light enhancement~\cite{li2021lowlightsurvey}, adverse weather restoration~\cite{liu2018desnownet,liu2019griddehazenet} and other degradation \cite{cao2022vdtr,bai2019single,zamir2022restormer,kupyn2019deblurgan,whang2022deblurring,zhang2020deblurring,liu2020densely,dabov2007image} independently. 
These methods made specific assumptions tailored to their respective problems. 
Consequently, a straightforward sequential amalgamation of these methods proves inadequate in addressing the concurrent degradation stemming from both low light and adverse weather.
Existing low-light enhancement methods~\cite{wang2020lightening,xu2022snr} primarily focus on intensity boosting, often treating other forms of distortion, such as rain, as commonplace elements. 
This approach often leads to the amplification of noise and blur in degraded regions after light enhancement.
Performing the weather or blur removal subsequent to light enhancement yields inconsistent outcomes.
When it comes to weather removal, nighttime weather condition images often contain saturated regions that do not conform to the blur that the model learned from daytime scenes. 
As a result, existing methods trained on datasets consisting only of daytime scenes cannot be seamlessly applied to the challenging nighttime images, as demonstrated in Fig.~\ref{fig:teaser_weather}.

To overcome the scarcity of real-world paired data encompassing both low-light and adverse conditions, we propose a data synthesis pipeline.
The pipeline involves modeling the degradation using classic imaging mathematical models.
The procedure unfolds through the sequential generation of adverse weather conditions followed by the imposition of low-light constraints.
After obtaining the data, we can incorporate both low-light enhancement and adverse condition restoration in a single context.
To address the joint degradation problem, the model needs to possess two fundamental capabilities.
First, the model should exhibit the capacity to discern varying levels of illumination. The substantial disparity in illumination levels~\cite{wu2022uretinex} between low-light and normal-light images presents a formidable obstacle in distinguishing underlying objects from the influence of adverse conditions. Complications like over-exposure and under-exposure further compound this task.
Second, the complex degradation problem poses a significant challenge due to its highly ill-posed nature. 
Therefore, the model is required to contain massive priors or generative power to infer missing information.


Recently, denoising diffusion probabilistic models have emerged as a promising paradigm in the realm of image generation, yielding impressive performance~\cite{ho2020denoising}.
During the inference process, diffusion models progressively denoise the input. 
This inherent progressive characteristic imbues the model with a substantial capacity to capture intricate texture details.
By treating weather degradation as intensive unconstrained noise characterized by diverse patterns, denoising networks can effectively suppress this noise and inpaint the image.
However, simple diffusion models still struggle to handle complex night scenes, as shown in the second row of Fig.~\ref{fig:teaser_weather}. 
The significant variation in light intensity renders the task difficult for models, \textit{e.g.}, WeatherDiff~\cite{ozdenizci2023restoring}, DiT~\cite{peebles2023scalable}, hindering effective learning and leading to issues such as color inconsistency, overexposure, and shading artifacts.




In this work, we propose a novel diffusion probabilistic model to handle this challenging task. 
In our approach, we employ a progressive denoising process driven by input image, where randomly sampled Gaussian noise is steadily removed using the devised network. To ensure structural coherence, these noise reduction stages are integrated into the diffusion process as contextual conditions.
To abridge the gap between low light and normal light conditions, we further include an illumination-guided module to guide the recovery process. 

The inclusion of an illumination prior aids the model in distinguishing between weather-induced degradation and the intended object, thereby facilitating an adaptive restoration of lighting levels. This leads to a more accurate recovery process.
Experiments demonstrate the effectiveness of our approach and its ability to handle different low-light image enhancement tasks.  Specifically, we validate the superiority of the proposed method across multiple datasets with large variances, which suggests our model could adapt easily to new scenarios and surpass state-of-the-art methods under various settings.

Our main contributions are presented as follows:

\begin{itemize}
    
    \item We propose a novel illumination-guided diffusion-based low-light enhancement method. Integrating the illumination module into the recovery process effectively mitigates issues related to color incongruity and under-/over-exposure.

    \item We introduce a meticulously crafted pipeline for synthesizing complex nighttime scene data. This inclusive framework encompasses degradation stemming from snow, rain, raindrops, haze, and fog.
    

\end{itemize}

%% file: tex/image_tex/weather/teaser_weather.tex
\begin{figure*}[t]
\begin{center}
\centerline{\includegraphics[width=1\linewidth]{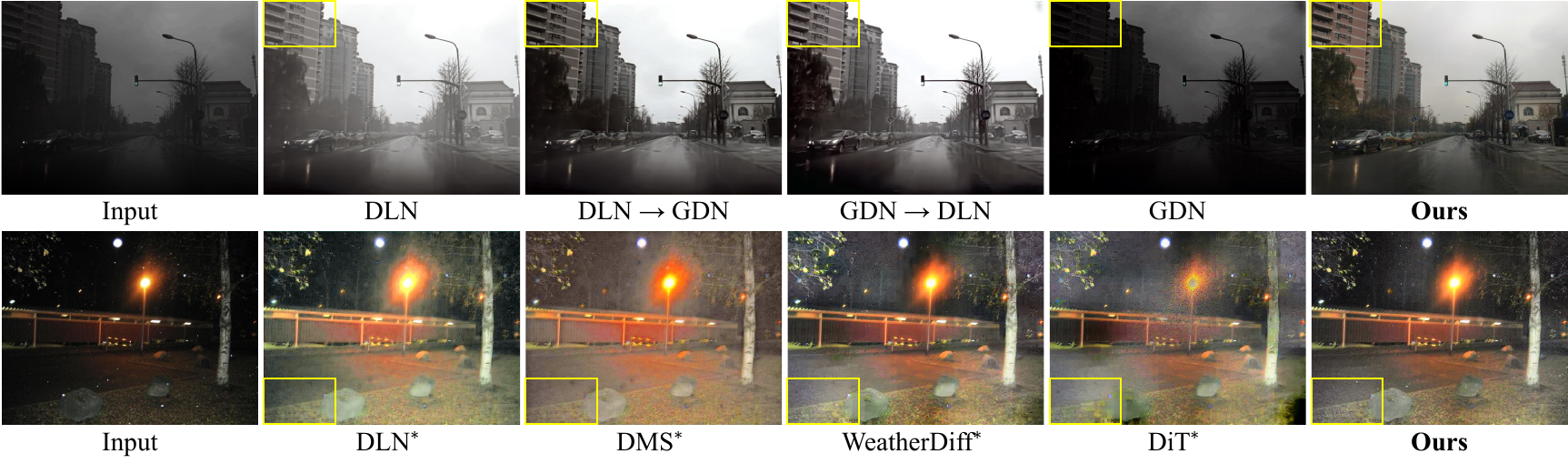}}
\caption{ 
\textit{\textbf{First row}}: A comparison reveals that existing low-light enhancement method DLN~\cite{wang2020lightening} and restoration method GDNet~\cite{liu2019griddehazenet} fail in dealing with complex night scene images.
A cascade of separate restoration methods cannot work harmoniously and may even amplify the degradation pattern, where $\to$ means applying the two methods sequentially. 
To address the problem, we synthesize diverse datasets for end-to-end training. 
\textit{\textbf{Second row}}: When applied to \textit{real-world images}, former methods retrained on our dataset still exhibit over-exposure problem, including DLN~\cite{wang2020lightening}, DDMSNet~\cite{zhang2021deep} and diffusion-based solutions (WeatherDiff~\cite{ozdenizci2023restoring} and DiT~\cite{peebles2023scalable}). In contrast, our approach yields satisfactory results with the help of illuminance guidance.} 
\label{fig:teaser_weather}
\vspace{-5pt}
\end{center}
\end{figure*}

%% file: main/2_relatedworks.tex
\section{Related Work}

\subsection{\textbf{Low-Light Image Restoration}}
Recently, Low-light image enhancement methods based on deep learning have achieved significant progress.
the efficacy of vision transformer-based architectures for image restoration has been a driving factor in this advancement.
Xu~\textit{et al}.~\cite{xu2022snr} proposed to adaptively enhance low-light images in a spatial-varying manner. 
Tu~\textit{et al.}~\cite{tu2022maxim} designed a general framework with a multi-axis gated MLP module for image processing. 
Cui~\textit{et al}.~\cite{cui2022you} built a lightweight transformer network to restore the normal light images from either low-light or over-exposure scenarios.
Jiang~\textit{et al}.~\cite{jiang2021enlightengan} used an unsupervised generative adversarial network.
Ma~\textit{et al}.~\cite{ma2022toward} presented a self-calibrated illumination learning module, which brightens images in real-world scenarios based on Retinex theory.
Retinex-based methods are also popular. 
For instance, 
Zhao~\textit{et al}.~\cite{zhao2021retinexdip} and Liang~\textit{et al}.~\cite{liang2022self} leveraged deep image prior to performing Retinex decomposition.
Guo~\textit{et al.}~\cite{guo2020zero} and Li~\textit{et al}.~\cite{li2021learning} developed an image-specific curve to approximate pixel-wise and higher-order curves.
While low-light enhancement methods concentrate primarily on enhancing illumination, they may inadvertently introduce artifacts when confronted with degradation that impacts the illumination. This unintended consequence arises from treating these degradations as integral to the enhancement objective.
In contrast, our method addresses this issue by considering the joint degradation conditions.

\subsection{\textbf{Adverse Weather Restoration}}
Previous works on weather recovery have focused on various aspects, such as snow removal, rain removal, raindrop removal, fog removal, and haze removal.
For snow removal, Zhang \textit{et al}. \cite{zhang2021deep} proposed a Deep Dense Multi-Scale Network (DDMSNet) that leverages semantic and geometric maps to remove snow. 
The network utilizes a self-attentive mechanism to learn semantic-aware and geometric-aware representations, resulting in clean images.
\cite{yasarla2020syn2real} proposed a semi-supervised learning framework for single-image de-raining using a Gaussian process. 
\cite{yasarla2020syn2real} allows the network to use synthetic datasets during training and promotes the use of unlabelled real-world data, improving the performance of rain removal.
\cite{quan2021removing} introduced a complementary cascaded network architecture for rain mark and raindrop removal. 
Their method removes raindrops first and then eliminates rain marks, followed by fusion using an attention based fusion module.
To address the challenges of fog and haze removal, \cite{jin2023structure} proposed a method that utilizes feature representations from a pre-trained visual transformer module. They also introduced uncertainty feedback learning and utilized the grayscale version of the input image to handle non-uniform haze regions and atmospheric light estimation.
%
\cite{liu2019griddehazenet} presented GridDehazeNet, a network for single-image dehazing. It consists of three modules: pre-processing, backbone, and post-processing. The pre-processing module generates diverse and relevant learning inputs, the backbone module employs an attention-based multiscale estimation grid network, and the post-processing module helps reduce artifacts in the output.
\textcolor{black}{Nevertheless, a significant portion of the previously mentioned weather recovery algorithms concentrates on daytime scenarios characterized by normal illumination, often neglecting the intricacies of low-light conditions encountered during nighttime.
Moreover, datasets and end-to-end algorithms tailored to tackle complex nighttime scenes remain absent.}

%% file: main/3_dataset.tex
\section{Our Data Synthesis Pipeline}



Efforts have been made to collect real-world paired data tailored for tasks like low-light enhancement~\cite{wei2018deep,yang2021sparse} and adverse condition restoration~\cite{liu2018desnownet,sakaridis2018semantic,fu2017removing,ren2019progressive,qian2018attentive,liu2021synthetic}.
However, the current landscape lacks datasets that encompass both of these scenarios concurrently.
The collection of such comprehensive datasets presents significant challenges, predominantly due to the inherent complexities of procuring paired samples and corresponding ground truth for complex night scenes.
Factors such as the dynamic nature of environmental conditions and the potential introduction of camera shake during data acquisition can lead to geometric and photometric misalignment, thereby further intensifying the intricacies of the data collection process.

To tackle this challenge, we propose a synthesis pipeline that facilitates the simultaneous modeling of both low-light and adverse conditions, as shown in Fig. \ref{fig:weather_pipeline}.
This approach enables us to generate datasets encompassing a diverse array of degradation scenarios within both domains.
Consequently, we procure a comprehensive collection of datasets, encompassing a total of 5 distinct categories. 
Tab.~\ref{tb:04_lol_weather_data} illustrates the sources of daytime scene data, wherein we apply diverse low-light and weather degradations to construct our dataset. 
We also provide the total number of training and testing pairs for reference.

\input{tex/table_tex/weather_blur/LOL-weather}

\subsection{\textbf{Adverse Weather Modeling}}
Adverse weather conditions wield a substantial influence over the fidelity and quality of images. Phenomena like raindrops, fog, or snow can induce blurring, diminished contrast, compromised visibility, and various other artifacts.
The modeling of adverse weather conditions under normal illumination can be achieved through the corresponding approaches below.

\textbf{Raindrop}~\cite{qian2018attentive} is modelled as follows:
\begin{equation}\label{eq:raindrop_model}
I=(1-M)\odot C + R ,
\end{equation}
where $I$ is the degraded image, $C$ is the clean image , $M$ is the mask and $R$ is the raindrop residual and $\odot$ denotes element-wise multiplication.

\textbf{Rain}~\cite{yang2016joint} with rain streaks is modeled as follows:
\begin{equation}\label{eq:rain_model}
I=T\odot(C + \Sigma_i^nR_i) + (1-T)\odot A,
\end{equation}
where $T$ is the transmission map produced by the scattering effect, 
$A$ is the atmospheric light in the scene, 
$i$ is the index of each rain streak, and $n$ is the total number of rain streaks.

\textbf{Snow}~\cite{liu2018desnownet} is generally modeled as follows:
\begin{equation}\label{eq:snow}
I =  (1-M)\odot C  + M \odot S,
\end{equation}
where $S$ corresponds to snowflakes.

\textbf{Fog}~\cite{jin2023structure} and \textbf{Haze}~\cite{li2017aod} can be modeled as:
\begin{equation}\label{eq:fog_haze_model}
I = T \odot C + (1-T) \odot A.
\end{equation}

Once the adverse weather conditions have been modeled, the subsequent task is to generate the corresponding adverse weather effect. 
This can be realized by sampling from the distributions that have been modeled,  thereby generating realistic raindrops, rain, fog, haze, or snow. 
These effects can then be combined with the normal image using automatic algorithms or image editing software, such as Photoshop.
By superimposing the generated adverse weather effects onto the normal-light image, we enable the creation of simulated adverse weather datasets that accurately reflect real-world scenarios.

\subsection{\textbf{Low Light Modeling}}
\label{sec:3.1}


Traditional low light modeling methods, such as Gamma correction, often introduce significant color deviation and noticeable warm tones.
Following LEDNet~\cite{zhou2022lednet}, we adopt a zero-shot low-light enhancement method Zero-DCE~\cite{guo2020zero} in reverse, EC-Zero-DCE,  to simulate light degradation. 
Specifically, we simulate the degradation of low light via a reversed curve adjustment in the exposure control loss. 

%
%

\input{tex/image_tex/weather/curve_eczerodce}
EC-Zero-DCE is used for estimating low-light conditions with controllable darkness levels by utilizing a modified exposure control loss. We introduce a random darkness parameter instead of a fixed exposure value. 
The process in EC-Zero-DCE can be formulated as follows: 
\begin{align}\label{EC}
EC_{n}(\mathbf{x, e}) &= EC_{n-1}(\mathbf{x, e}) \nonumber \\
&\quad + \mathcal{M}_{n}(\mathbf{x, e})EC_{n-1}(\mathbf{x, e})(1-EC_{n-1}(\mathbf{x, e})),
\end{align}
where $x$ is the input image, $EC_{0}=x$, we iterate it with high-order for more powerful adjustment capability according to Zero-DCE~\cite{guo2020zero}. $\mathcal{M}$ is a learnable pixel-wise adjustment parameter map with the same size as the given image. Benefiting from the property of pixel-wise adjustment of $\mathcal{M}$ for illumination in EC-Zero-DCE, each pixel of the given input image can be darkened dynamically, which can simulate real-world darkness with weather or other degradation conditions. Therefore, we can leverage this property to set the lightness for the objects such as background and rain  which have different transparency. $e$ is an exposure parameter that can control the degree of darkening.  
We set $n=10$, $e \in [0.05, 0.3]$ as a random value to increase the diversity of darkness levels and the image will be darker when $e$ is lower. As shown in Fig.\ref{fig:curve_eczerodce}, the EC-Zero-DCE curve maintains monotonicity, which keeps the consistency of darkening tendency in real-world scenes. 

It is pertinent to emphasize that our approach executes pixel-wise and spatially variant light adjustments, as opposed to uniform light degradation. This distinctive method results in the production of low-light images that possess enhanced naturalism and realism, thereby circumventing undesirable color artifacts.

\subsection{\textbf{Simulating Adverse Weather in Low-Light Conditions}}

In real-world scenarios, taking a night snow scene as an example, the color of snowflakes is influenced by ambient light, often appearing as a shade of greyish-white.
If the dataset construction entails the initial darkening of the scene followed by the subsequent addition of snowflakes, there is a risk of producing overly luminous snowflake colors that deviate from the authentic luminance of the nighttime environment.
Therefore, we opt to introduce adverse weather conditions prior to the subsequent darkening process, as depicted in Fig.~\ref{fig:weather_pipeline}. This approach adeptly harmonizes the brightness of the degradation patterns within the dimly lit environment.

This dataset offers diverse examples of adverse weather conditions and low-light scenarios.
This comprehensive dataset equips our model with the capacity to learn and generalize adeptly across a spectrum of distinct situations. 
The amalgamation of these two categories of degradations engenders a dataset that is notably more complex and realistic, serving as a training and evaluation resource.

\input{tex/image_tex/weather/pipeline}

%% file: tex/table_tex/weather_blur/LOL-weather.tex
 \begin{table}[ht]
\centering
\renewcommand{\arraystretch}{1.3}
\caption{Details of dataset composition. `Training' and `Testing' refer to the pair number of corresponding data partition.}
\label{tb:04_lol_weather_data}
\resizebox{\linewidth}{!}{
\begin{tabular}{@{} l |c|c|c @{}}
\hline
 & Daytime scene data sources & Training & Testing \\
\hline 
LOL-Fog& CityscapeFog~\cite{sakaridis2018semantic} & 2975&1525\\
LOL-Haze& Haze4K~\cite{fu2017removing}& 3000 & 1000 \\
LOL-Rain& Rain12600~\cite{fu2017removing}, Rain1400~\cite{ren2019progressive}& 900 & 100\\
LOL-Raindrop & Raindrop~\cite{qian2018attentive}& 861 & 58\\
LOL-Snow & Snow100K~\cite{liu2018desnownet} & 1000 & 100\\
\hline
\end{tabular}
}
\end{table}

%% file: tex/image_tex/weather/curve_eczerodce.tex
\begin{figure}[t]
\centering 
\includegraphics[width=0.8\linewidth]{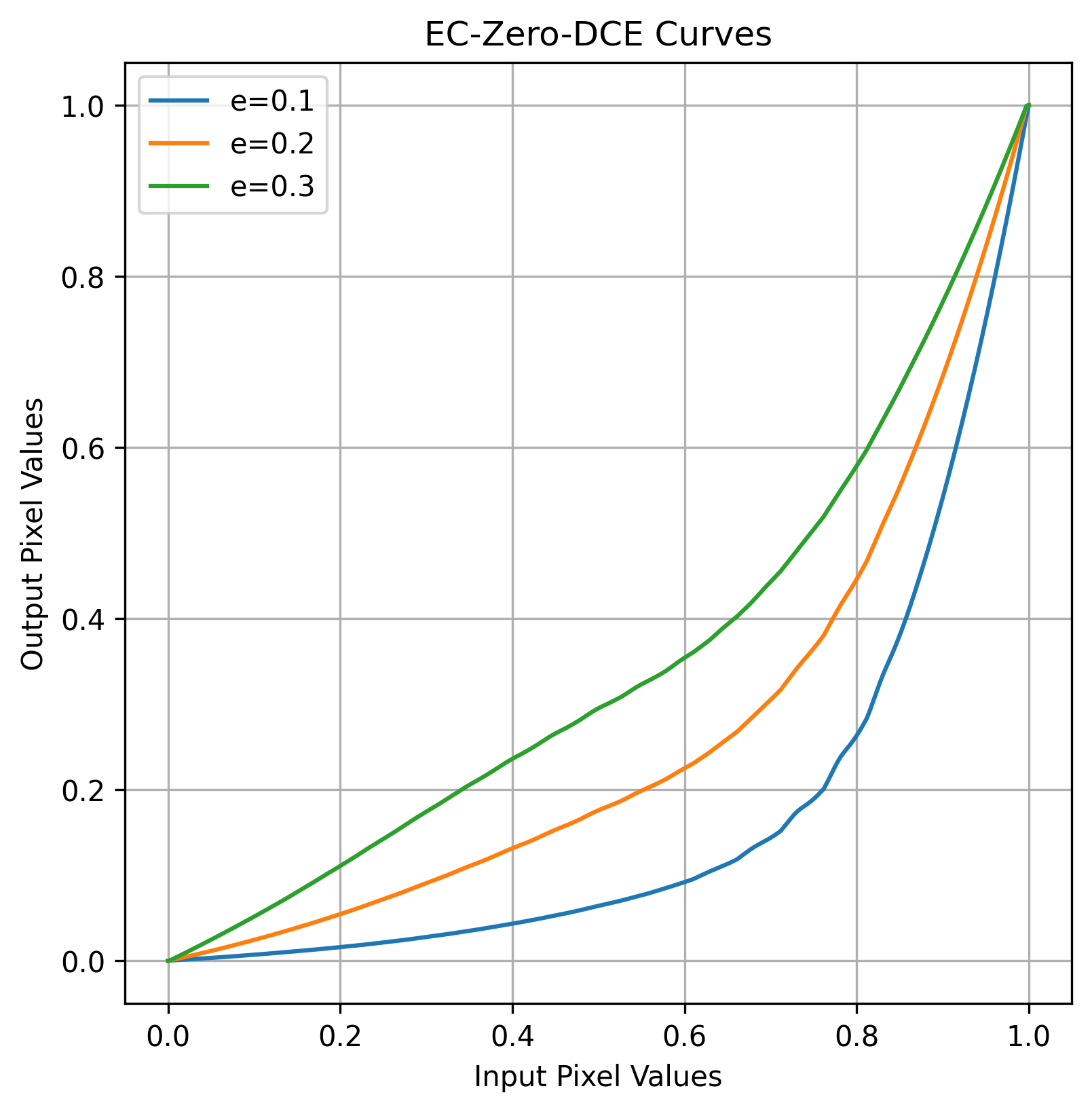}
\caption{EC-Zero-DCE Curves with different adjustment parameters exposure $e$. The horizontal axis represents the input pixel values while the vertical axis represents the output pixel values.}
\label{fig:curve_eczerodce}
\end{figure}

%% file: tex/image_tex/weather/pipeline.tex
\begin{figure}[t]
\centering 
\includegraphics[width=1\linewidth]{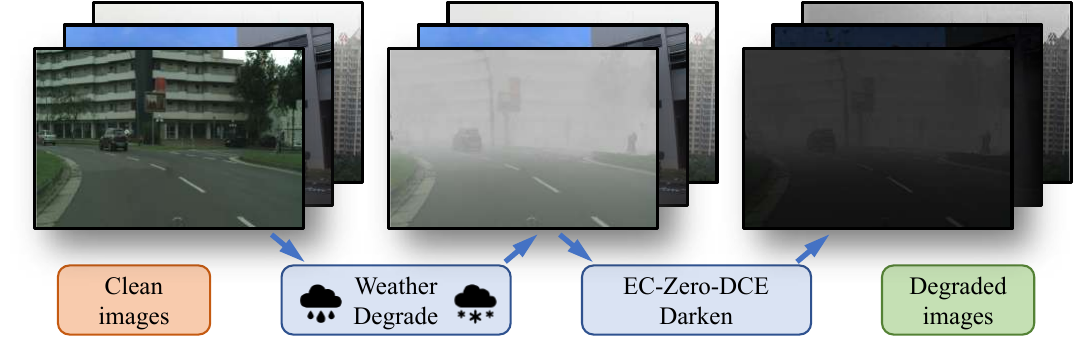}
\captionsetup{justification=centering}
\caption{An overview of our data synthesis pipeline.}
\label{fig:weather_pipeline}
\end{figure}

%% file: main/3_proposedmethod.tex
\section{Our Methodology}
\input{tex/image_tex/low-light/method}

\noindent \textbf{Illumination Estimation}.
%
Prior methods employ Retinex theory as a formula for recovery. 
However, the ideal assumption that treats the reflection component as the enhanced result does not always hold, especially when confronted with diverse illumination characteristics. This can result in unrealistic enhancements marked by the loss of details and distorted colors. 
Additionally, noise may remain or be amplified in the enhanced results due to the vacancy of modeling noise.
Hence, we opt to use the illumination map as our restoration guidance.
To address these shortcomings, we employ an illumination map as the guiding principle for restoration. 

In this context, we make use of a \textit{pre-trained} illumination estimation model $E_{sci}$~\cite{ma2022toward}. 


$E_{sci}$ builds upon Retinex theory, which  models the low-light degraded image $y$ as $y = z \odot x$, where $z$ denotes the clean image and $x$ represents the illumination component. To effectively estimate illumination, $E_{sci}$ adopts a progressive refinement strategy through a cascade illumination calibration network. The estimation process is formulated as: 
\begin{equation}
    x_{t+1}=x_{t}+H_{\theta}(x_t),
\end{equation}
where $x_t$ represents the illumination estimation at the $t$-th stage, with the initial state $x_0$ set to the input low-light image $y$. Instead of directly learning the mapping from low-light images to illumination, this residual learning approach leverages the prior knowledge that illumination patterns generally share high similarities with low-light observations in most regions, which improves the estimation robustness. 
The illumination map $x_{illu}$ obtained from $E_{sci}$ is then used as a prior to guide the recovery process.

\input{tex/image_tex/weather_blur/lolweather}
\input{tex/image_tex/weather/04_deweather}
\vspace{2pt}
\noindent \textbf{Illumination-Guided Diffusion Model}.
Unlike image generation tasks, image restoration task requires the restored images to be consistent with given degraded inputs, including the spatial structure and illumination contrast.
Given the low-light images $\widetilde{x}$ and the extracted illumination $x_{illu}$, we learn a conditional diffusion model. 
The diffusion model defines a forward process that gradually adds Gaussian noise to the data, and a backward process that learns to denoise. The model can be trained by optimizing:
\begin{equation}\label{eq:obj}
\mathbb{E}{x_0,t,\epsilon_t \sim \mathcal{N}(0, \mathbf{I})} \left [ \lVert \epsilon_t - \epsilon\theta(x_t, t) \rVert ^2 \right ],
\end{equation}
where $\epsilon_t$ is the noise added at timestep $t$, and $\epsilon_\theta$ is a neural network that predicts this noise.
Following \cite{song2020ddim}, we formulate the implicit sampling forward process as follows:
\vspace{-10pt}
\begin{equation}\label{eq:cond_model}
p_\theta(x_{0:T}|\widetilde{x},x_{illu}) = p(x_T) \prod_{t=1}^T p_\theta(x_{t-1}|x_t, \widetilde{x},x_{illu}).
\end{equation}
Then, we use $\epsilon_\theta(x_t, \widetilde{x},x_{illu},t)$ as the noise estimation network when optimizing Eq.~\eqref{eq:obj}.
The network structure of $\epsilon_\theta(x_t, \widetilde{x},x_{illu},t)$ is shown in Fig.~\ref{fig:pipeline}. We employ two injection pathways to maintain the spatial structure and restoration intensity separately.
For spatial structure, we concatenate latent $x_t$ and low-light images $\widetilde{x}$ along the channel dimension before feeding them into the network. 

To achieve adaptive image enhancement while circumventing issues like over-exposure or under-exposure, we leverage an illumination map as a guiding principle for restoration, which directs the network's attention toward low-light regions.
Additionally, the integration of the illumination map into our approach serves to enable the model to synthesize harmonious outcomes by harnessing non-local information and achieving a more comprehensive contextual perception.
We introduce an illumination guidance encoder $E_{illu}$ that projects illumination map $x_{illu}$ to an intermediate pyramid features $[F_0, F_1, \cdots, F_m]$.
The features $F_i \in \mathbb{R}^{H_i \times W_i \times C_i}$ are then mapped to the intermediate layers $L_i \in \mathbb{R}^{H_i \times W_i \times C_i}$ of the UNet accordingly via a cross-attention layer implementing Attention($Q$, $K$, $V$)=softmax$(\frac{QK^T}{\sqrt{d}})\cdot V$, with
\begin{equation}
Q=W_Q^i\cdot \phi(F_i), K=W_K^i\cdot \phi(L_i), V=W_V^i\cdot \phi(L_i),
\end{equation}
where $\phi(\cdot)$ denotes a flatten function and $W_Q^i, W_K^i, W_V^i$ are learnable projection matrices.
The features residing within distinct layers are engaged in a synergistic manner to attain a more refined and expansive perceptual field.

\vspace{2pt}
\noindent \textbf{Patch-Based Restoration}.
To accommodate the diverse dimensions of real-world images, we adopt a patch-based strategy for the restoration process.
During the training phase, we extract fixed-size patches from both low-light images and their corresponding positions in the ground truth data. Simultaneously, we extract the illumination map from the original image and tailor it to match the dimensions of the relevant area.
During the inference stage, we decompose the images into patches, applying light enhancement individually to each patch before seamlessly blending them back into their original dimensions. However, given that local patches lack adjacency information, directly merging patches from separately enhanced results can yield unsightly edge artifacts.
Inspired by \cite{bar2023multidiffusion}, we adopt a technique wherein overlapping patches are sampled. This enables the smoothing of the restored illumination in shared regions during the reverse sampling process.
In each time step $t$, we employ the average estimated noise of each pixel as the latent updating direction. 
Specifically, we calculate the accumulation of the predicted noise $\epsilon_t^i$ based on their respective patch location $i$. 
This accumulated noise is subsequently normalized by the number of pixel counts. Through this process, we ascertain the latent from the previous time step. 

%% file: tex/image_tex/low-light/method.tex

\begin{figure}[t]
\begin{center}
    \centerline{\includegraphics[width=1\linewidth]{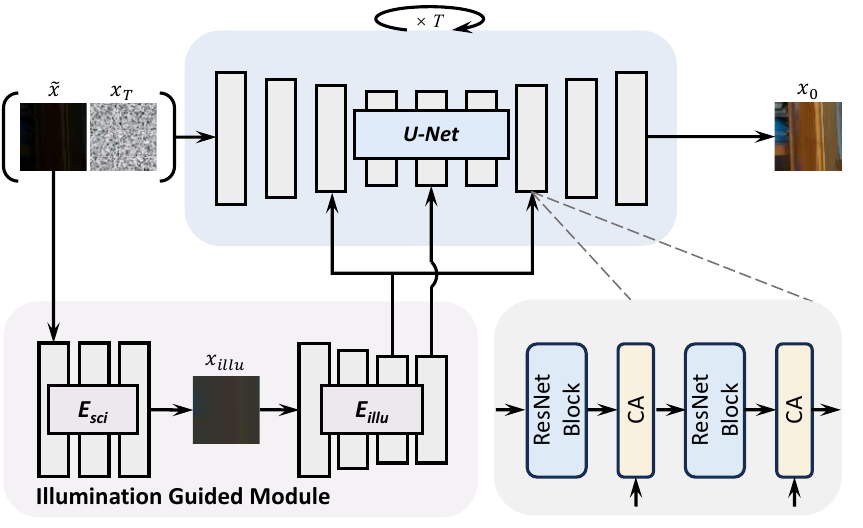}}
\caption{Our framework for low-light image enhancement. We first estimate an illumination map for guiding the restoration process. The illumination map can help prevent over-exposure and under-exposure problems.}
\label{fig:pipeline}
\vspace{-0.5cm}
\end{center}
\end{figure}

%% file: tex/image_tex/weather_blur/lolweather.tex
  


\begin{figure*}[t]
\begin{center}
    \centerline{\includegraphics[width=1\linewidth]{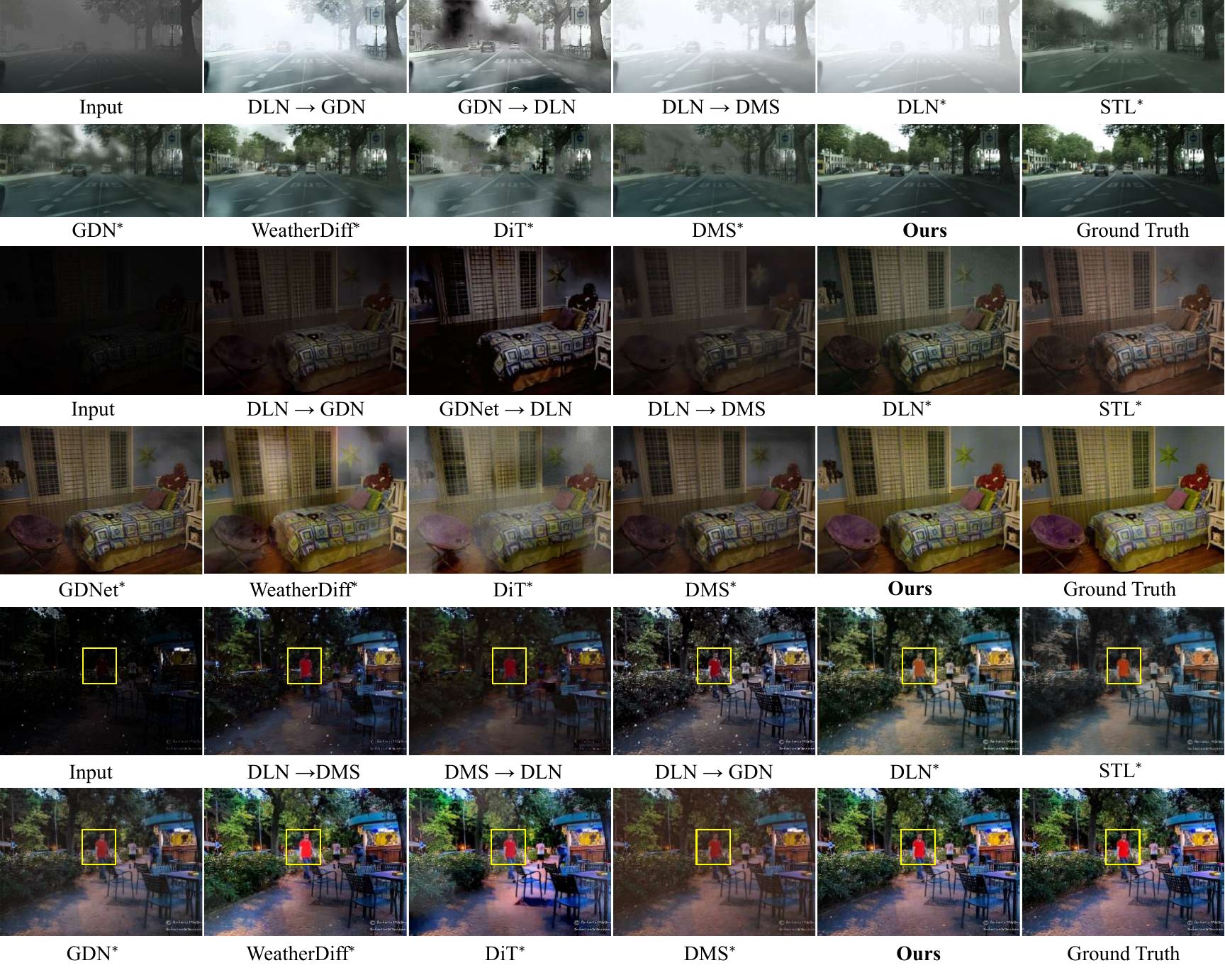}}
\caption{Our framework for low-light image enhancement. We first estimate an illumination map for guiding the restoration process. The illumination map can help prevent over-exposure and under-exposure problems.}
\label{fig:04_lol_weather_supp}
\vspace{-0.5cm}
\end{center}
\end{figure*}








%% file: tex/image_tex/weather/04_deweather.tex
\begin{figure*}[ht]
  \subfloat[LOL-Rain]{\includegraphics[width=\textwidth]{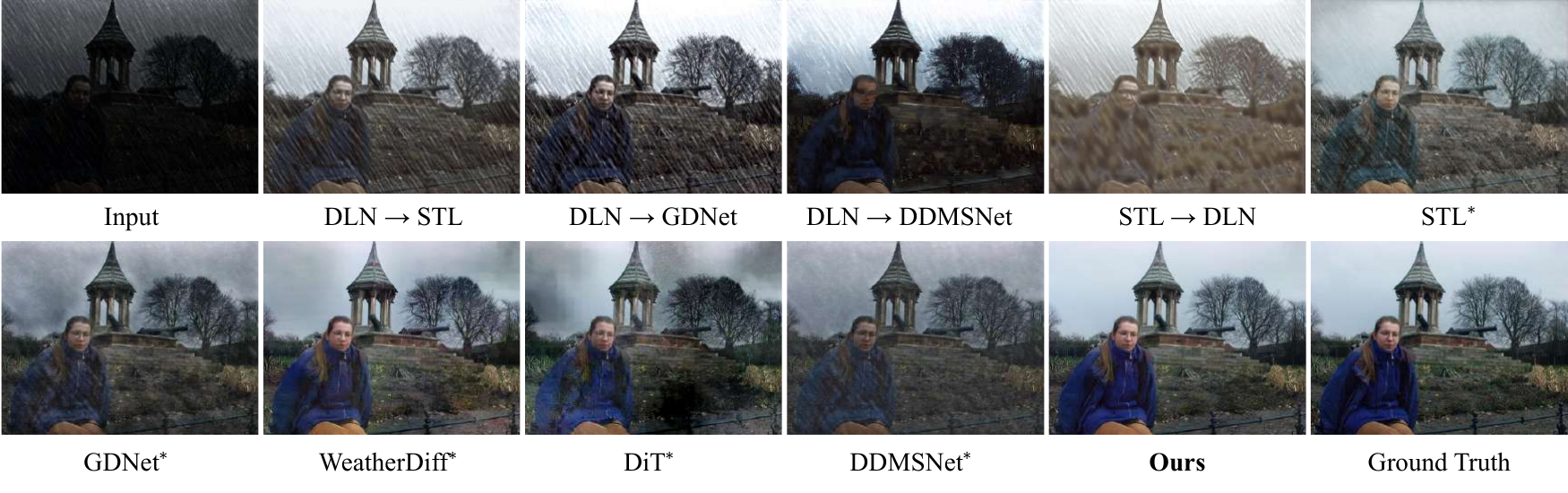}} \qquad
  \subfloat[LOL-Raindrop]{\includegraphics[width=\textwidth]{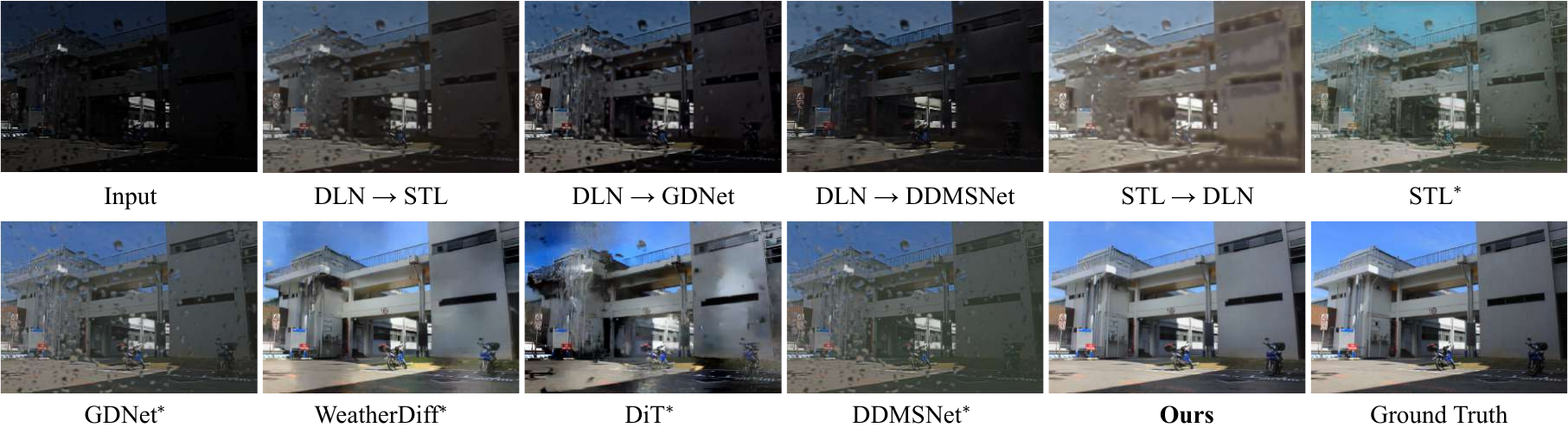}}
  \caption{Qualitative comparison on LOL-Rain and LOL-Raindrop.}
  \label{fig:04_lol_rain_lol_raindrop}
\end{figure*}





%% file: main/4_experiment.tex
\input{tex/table_tex/weather/deweather}

\input{tex/table_tex/low-light/lol_v1}

\input{tex/table_tex/deblur/lol_blur_noise}

\section{Experiments}
\subsection{\textbf{Experimental Settings}}

\noindent \textbf{Datasets}.
We synthesize low-light adverse weather data by augmenting the original scene dataset~\cite{liu2018desnownet,sakaridis2018semantic,fu2017removing,ren2019progressive,qian2018attentive,liu2021synthetic} with various weather conditions, including LOL-rain, LOL-raindrop, LOL-snow, LOL-fog, and LOL-haze. 
We also validate our methods on the publicly available LOL-blur-denoise dataset~\cite{zhou2022lednet} that contains the joint task of low-light enhancement, deblurring, and denoising, which closely resembles our joint restoration settings. 


\vspace{2pt}
\noindent \textbf{Comparison Methods}. 
For complex night scene restoration, we compare our methods with $12$ baselines. Given the relatively nascent nature of complex night scene restoration, making direct head-to-head comparisons with pre-existing methods is not feasible. Therefore, we select a set of representative low-light enhancement methods, including DLN~\cite{wang2020lightening}, and weather restoration methods, including STL~\cite{yasarla2020syn2real}, GDN~\cite{liu2019griddehazenet}, and DMS~\cite{zhang2021deep}, which we combined and adapted for evaluation.
We also include two diffusion-based solutions, \textit{i.e.}, WeatherDiff~\cite{ozdenizci2023restoring} and DiT~\cite{peebles2023scalable}, to validate the effectiveness of our model.
We categorize the combinations of the methods into three groups: \texttt{Enhancement$\to$Restoration}, \texttt{Restoration$\to$Enhancement} and \texttt{End-to-end training}. 
We use their released models for the first two configurations. We retrain the models on our datasets under the same \texttt{End-to-end training} setting.


\noindent \textbf{Implementation Details}
We set the diffusion step as $1,000$ and the sampling time step as $40$. 
During training, we randomly sample a time step $T$ from the range of the diffusion step.
We trained our models on the patch size of $64 \times 64$.
Following SCI~\cite{ma2022toward}, illumination estimator $E_{sci}$ are pre-trained on datasets Darkface~\cite{yang2020advancing} in an unsupervised manner. Notably, there is no scene overlap between the training datasets and our low-light adverse weather dataset. The unsupervised training approach enables SCI~\cite{ma2022toward} to effectively address a wide range of light-degraded scenarios.
Illuminations are injected via cross attention at $16 \times 16$ and $8 \times 8$ feature map resolution. 
To simulate low-light degradation, we employ EC-Zero-DCE pre-trained on low-light datasets~\cite{zhou2022lednet}, which do not include any of the datasets we use for synthesizing datasets in this paper.
We used an Adam optimizer with a fixed learning rate of $0.00002$ without weight decay. An exponential moving average with a weight of $0.999$ was applied during parameter updates.
During inference, the grid step size is set as $16$.
\textcolor{black}{All experiments are conducted on a single NVIDIA GeForce RTX 4090Ti GPU.}

\input{tex/image_tex/weather_blur/lol_blur_noise}
\input{tex/image_tex/weather_blur/realhaze}

\subsection{\textbf{Evaluation on Complex Night Scenes}}
\noindent \textbf{Quantitative Comparison}.
Tab.~\ref{tb:04_lol_weather} presents the quantitative results on the LOL-weather dataset. 
In all settings, our method outperforms all baseline methods.
The two-stage methods perform poorly as they make independent assumptions and neglect other degraded factors. 
\texttt{Restoration$\to$Enhancement} methods perform even worse, mainly because the original adverse weather dataset only contains daytime scenes. When these methods are directly applied to our dataset, they struggle to detect weather artifacts, and subsequent light enhancement may even mistakenly treat the weather artifacts as light effects, leading to worsened results.
\texttt{Enhancement$\to$Restoration} methods achieve a relatively higher SSIM because of their initial enhancement of lightening, which makes the images more similar in the input domain.
On the other hand, \texttt{End-to-end training} methods all achieve relatively higher performance, with an average PSNR over $19.00$. 

Our proposed model demonstrates versatility in addressing a range of low-light degraded complicated scenes. Our network excels in eliminating noise, even in diverse and irregular patterns. The illumination-guided module can navigate the illumination restoration process. 
In addition to CNN-based and GAN-based methods, we also conduct comparisons with diffusion-based restoration methods like Palette~\cite{saharia2022palette,ozdenizci2023restoring} and transformer-based diffusion methods like DiT~\cite{peebles2023scalable}. Apart from one-task models, we conduct experiments with multi-task models, such as MIR~
\cite{zamir2020learning} and WeatherDiff~\cite{ozdenizci2023restoring}. Among all the methods, our method stands out with a significant improvement, showcasing its superior performance over both multi-task and single-task models. 

\input{tex/image_tex/low-light/lol_v1}

\vspace{2pt}
\noindent \textbf{Qualitative Comparison}. 
The qualitative results of LOL-Fog, LOL-Haze, and LOL-Snow are shown in Fig.~\ref{fig:04_lol_weather_supp}. 
The qualitative results of LOL-Rain and LOL-Raindrop are presented in Fig.~\ref{fig:04_lol_rain_lol_raindrop}. 
The qualitative outcomes highlight the effectiveness of our approach.
which aligns well with our quantitative evaluations. 


To demonstrate the generalizability of our dataset and network in the wild, we also test on more real-world complex night scene images.
These real-world images are sourced from datasets such as RESIDE~\cite{li2018benchmarking} (Real haze night subset), and snow100k~\cite{liu2018desnownet} (Real snow night subset). As shown in Fig.~\ref{fig:04_lol_real_haze}, 
our method exhibits robust performance in real-world scenarios, which not only validates the effectiveness of our dataset but also underscores the model's capability to handle diverse night weather patterns. Note that none of the models have been exposed to these real images.



\subsection{\textbf{Evaluation on Multiple Degradation Dark Scenarios}}
In addition to the constructed complex weather scenes in our dataset, the potential for multiple combinations of night degradation scenarios is vast, leading to increased complexity and challenge. 
To address this, we conducted an experiment using the LOL-Blur-Noise dataset~\cite{zhou2022lednet}, which comprises pairs of images with varying levels of darkness, blur, and noise across different scenarios. 
Visual comparisons are presented in Fig.~\ref{fig:04_lol_blur_noise} and quantitative comparisons are shown in Tab.~\ref{tb:lol_blur_noise}.
Our approach excels at managing complex night scenes, even when they entail multiple combinations of degradation, demonstrating the versatility of our model across diverse scenarios.

\subsection{\textbf{Evaluation on Simple Low Light Scenarios}}

To further validate the effectiveness of our framework, we conducted experiments on pure low-light image datasets. 
The comparisons on LOL-v1 is shown in Tab.~\ref{tb:lol_v1}.
Our method outperforms all baseline approaches. 
Since the illumination-guided networks can effectively perceive the original scene, our framework can enhance the images and restore the illumination adaptively. The qualitative results are shown in Fig.~\ref{fig:lol_v1_new}

\input{tex/ablation/ablation_grid}

\vspace{-5pt}
\subsection{\textbf{Ablation Study}}

\noindent \textbf{Illumination-Guided Module}. 
To prevent over-exposure problems, we introduce an illumination-guided module to assist the restoration process. In Tab.~\ref{tb:illu}, we can find that
the proposed illumination guidance module can improve performance consistently in various datasets.

\vspace{2pt}
\noindent \textbf{Grid Step Selection}. 
We propose a patch-based low-light enhancement strategy for various input sizes. 
We slide the enhancement window with a fixed grid step. 
We perform an ablation study on the selection of grid step on the LOL-V1 dataset.
As shown in Tab.~\ref{tb:ablation_grid}, when the grid size reached $16$, the performance gradually converged. 
We set the grid step as $16$ as our hyper-parameter for the trade-off.
We further visualize the results in Fig.~\ref{fig:grid}; when the grid step increases larger than $32$, there exist differences in color and exposure between blocks. In the meantime, black borders also appear along the edges of the reconstructed image due to the grid size may not evenly divide the image dimensions when it is over 32.

\input{tex/table_tex/ablation/illumination}
\input{tex/table_tex/ablation/grid}

%% file: tex/table_tex/weather/deweather.tex
\begin{table*}[t]
\centering
\scriptsize
\setlength{\tabcolsep}{2pt}
\renewcommand{\arraystretch}{1.5} 
\caption{
Quantitative comparison on low-light adverse weather dataset. `*' denotes the network is retrained on our dataset.}
\label{tb:04_lol_weather}
\resizebox{\textwidth}{!}{
\begin{tabular}{c|c|ccc|ccc|ccccccccc}
\hline   
 &  & \multicolumn{6}{c|}{\textbf{Two-stage Methods}} &  \multicolumn{9}{c}{\textbf{One-stage Methods (End-to-End)}} \\
\hline
\multirow{2}{*}{Dataset} & Enhancement 
& DLN  & DLN  & DLN  & STL & GDN & DMS  
& \multirow{2}{*}{STL*} & \multirow{2}{*}{GDN*} & \multirow{2}{*}{DMS*}  & \multirow{2}{*}{DLN*}   & \multirow{2}{*}{MIR*} & Weather & \multirow{2}{*}{Palette*}  & \multirow{2}{*}{DiT*}  & \multirow{2}{*}{Ours} \\
 & Restoration & STL & GDN & DMS & DLN & DLN & DLN 
 &  &  &  &  &  & Diff* & & & \\ 
\hline
\multirow{2}{*}{LOL-Fog} & PSNR & 9.44 & 10.74 & 10.55 & 8.85 & 11.47 & 11.86 & 20.15 & 19.83 & 17.35 & 17.61 & 17.12 & 20.38 & 16.58 & 16.86 &  \textbf{29.34} \\
 & SSIM & 0.541 & 0.608 & 0.586 & 0.541 & 0.710 & 0.677 & 0.807 & 0.841 & 0.804 & 0.835 & 0.783 & 0.893 & 0.669 & 0.705 &  \textbf{0.956} \\ 
\hline
\multirow{2}{*}{LOL-Haze} & PSNR & 15.86 & 16.87 & 14.91 & 16.22 & 18.93 & 15.79 & 20.15 & 18.72 & 17.78 & 20.42 & 17.88 & 17.85 & 16.99 & 16.33 &  \textbf{27.74} \\
 & SSIM & 0.759 & 0.782 & 0.680 & 0.761 & 0.881 & 0.755 & 0.857 & 0.853 & 0.834 & 0.912 & 0.821 & 0.901 & 0.770 & 0.691 &  \textbf{0.981} \\ 
\hline
\multirow{2}{*}{LOL-Rain} & PSNR & 18.88 & 18.08 & 15.72 & 19.08 & 19.53 & 16.14 & 19.32 & 20.31 & 18.16 & 21.43 & 19.00 & 23.76 & 18.91 & 16.94 &  \textbf{27.54} \\
 & SSIM & 0.669 & 0.616 & 0.619 & 0.672 & 0.642 & 0.647 & 0.723 & 0.778 & 0.732 & 0.808 & 0.696 & 0.873 & 0.630 & 0.610 &  \textbf{0.903} \\ 
\hline
\multirow{2}{*}{LOL-Raindrop} & PSNR & 18.98 & 17.60 & 16.36 & 17.68 & 18.03 & 18.30 & 19.32 & 17.54 & 17.26 & 20.48 & 20.00 & 19.26 & 18.09 & 17.57 &\textbf{26.03} \\
 & SSIM & 0.788 & 0.724 & 0.671 & 0.707 & 0.784 & 0.738 & 0.767 & 0.752 & 0.735 & 0.851 & 0.784 & 0.848 & 0.638 & 0.690 &  \textbf{0.922} \\ 
\hline
\multirow{2}{*}{LOL-Snow} & PSNR & 19.32 & 18.28 & 14.82 & 19.34 & 19.46 & 14.21 & 20.52 & 18.59 & 17.54 & 23.42 & 19.78 & 21.82 & 16.12 & 18.35 &  \textbf{27.50} \\
 & SSIM & 0.774 & 0.738 & 0.606 & 0.709 & 0.769 & 0.580 & 0.803 & 0.800 & 0.771 & 0.861 & 0.741 & 0.898 & 0.563 & 0.739 &  \textbf{0.938} \\ 
\hline
\end{tabular}
}
\vspace{-0.3cm}
\end{table*}

%% file: tex/table_tex/low-light/lol_v1.tex
\begin{table*}
\centering
\caption{Quantitative comparison on LOL-V1}
\small
\label{tb:lol_v1}
\resizebox{\linewidth}{!}{
\begin{tabular}{@{} l cccc cccc cccc c @{}}
\hline
Method & Dong &  LIME &  MF &  SRIE &  BIMEF &  DRD &  RRM &  SID &  DeepUPE &  KIND & DeepLPF & FIDE & WeatherDiff \\ 
PSNR & 16.72 & 16.76 & 18.79 & 11.86 & 13.86 & 16.77 & 13.88 & 14.35 & 14.38 & 20.87 & 15.28 & 18.27 &24.03 \\
SSIM & 0.580 & 0.560 & 0.640 & 0.500 & 0.580 & 0.560 & 0.660 & 0.436 & 0.446 & 0.800 & 0.473 & 0.665 &0.918  \\
\hline
Method & LPNet &  MIR-Net &  RF &  3DLUT &  A3DLUT &  Band &  EG &  Retinex &  SGM & IPT &  SNR & DiT  &  \textbf{Ours} \\ 
PSNR & 21.46 & 24.14 & 15.23 & 14.35 & 14.77 & 20.13 & 17.48 & 18.23 & 17.20 & 16.27 & 24.61 &16.01   &\textbf{25.07} \\
SSIM & 0.802 & 0.830 & 0.452 & 0.445 & 0.458 & 0.830 & 0.650 & 0.720 & 0.640 & 0.504 & 0.842 &0.761 &\textbf{0.921} \\
\hline
\end{tabular}
}
\end{table*}

%% file: tex/table_tex/deblur/lol_blur_noise.tex
\begin{table}
\centering
\caption{Comparison on LOL-Blur-Noise dataset}
\label{tb:lol_blur_noise}
\small
\begin{tabular}{@{} l cccc@{}}
\hline
Method & KinD++ & DRBN & DeblurGAN-v2 & DMPHN \\
PSNR &  21.26   &  21.78   &  22.30   &  22.20   \\
SSIM &  0.75   &  0.77  &  0.75   &  0.82   \\ 
\hline
Method & MIMO & LEDNet & WeatherDiff  & \textbf{Ours} \\
PSNR &  22.41 & \textbf{25.74} & 21.03 &24.62 \\
SSIM &  0.84 & 0.85 & 0.80 &\textbf{0.87} \\
\hline
\end{tabular}
\end{table}

%% file: tex/image_tex/weather_blur/lol_blur_noise.tex
\begin{figure*}[t]
\begin{center}
\centerline{\includegraphics[width=\linewidth]{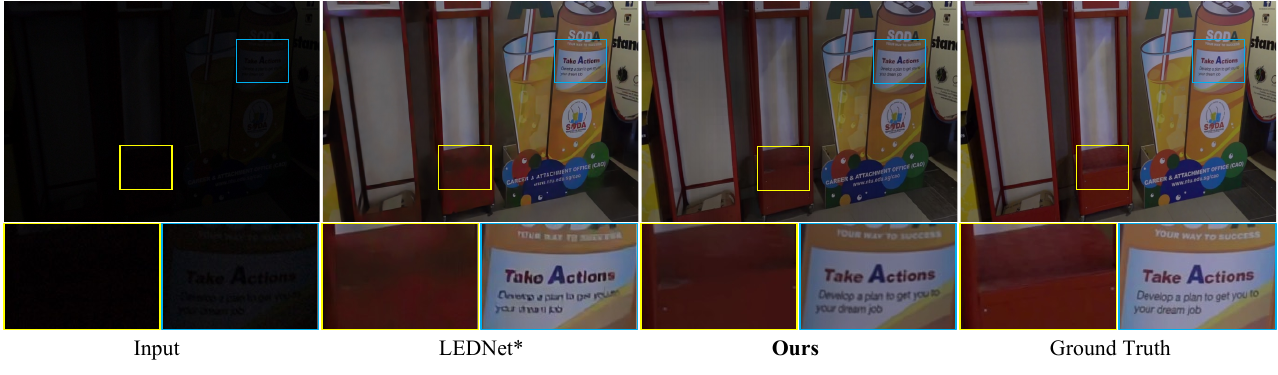}}
\vspace{-2pt}
\caption{Qualitative comparison on LOL-Blur-Noise. }
\label{fig:04_lol_blur_noise}
\end{center}
\vspace{-25pt}
\end{figure*}

%% file: tex/image_tex/weather_blur/realhaze.tex
\begin{figure}[t]
\begin{center}
\centerline{\includegraphics[width=\linewidth]{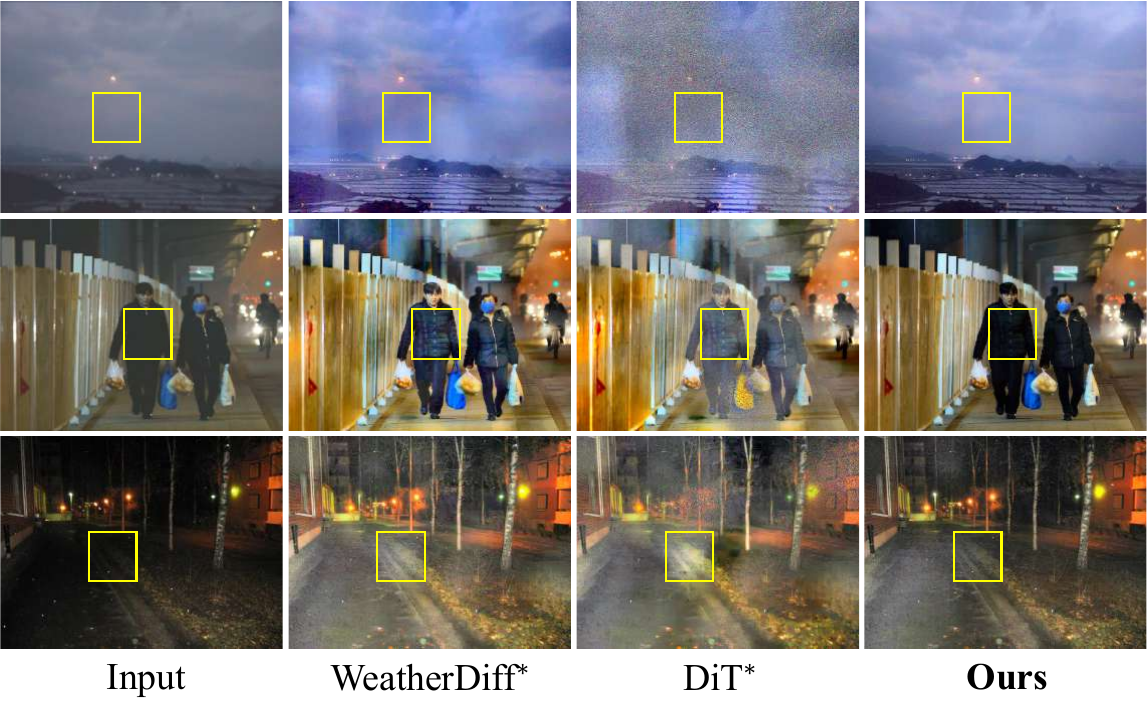}}
\vspace{-2pt}
\caption{ Qualitative comparison on real snowy images from Snow100K dataset. 
Note all models were trained on our synthetic dataset for a fair comparison. 
}
\vspace{-10pt}
\label{fig:04_lol_real_haze}
\end{center}
\end{figure}

%% file: tex/image_tex/low-light/lol_v1.tex
\begin{figure}[t]
\begin{center}
\centerline{\includegraphics[width=1\linewidth]{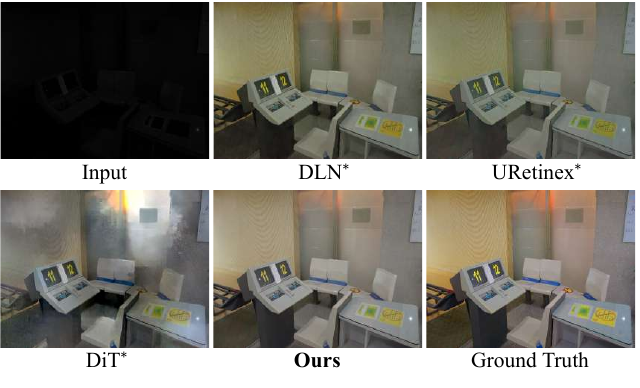}}
\caption{Qualitative comparison on LOL-V1.}
\label{fig:lol_v1_new}
\end{center}
\end{figure}

%% file: tex/ablation/ablation_grid.tex
\begin{figure}[t]
\begin{center}
\centerline{\includegraphics[width=\linewidth]{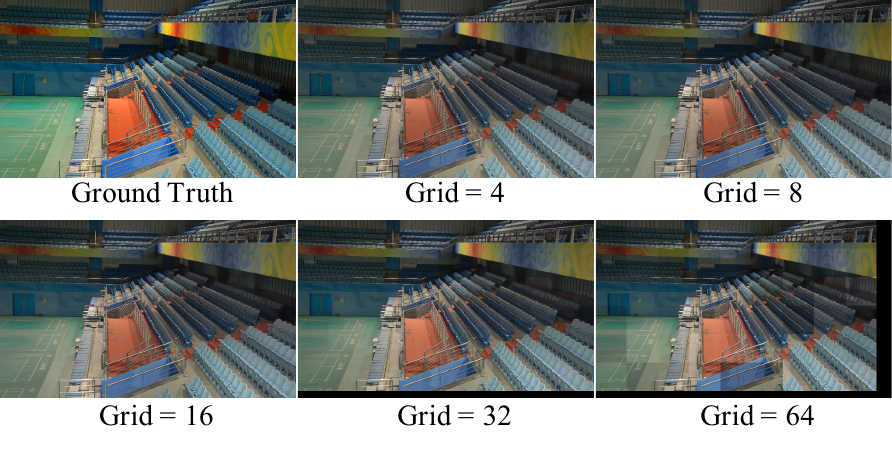}}
\caption{Ablation on grid step selection of LOL-V1 dataset}
\label{fig:grid}
\vspace{-10pt}
\end{center}
\end{figure}

%% file: tex/table_tex/ablation/illumination.tex
\begin{table}[t]
\centering
\renewcommand{\arraystretch}{1.5} 
\small
\caption{Ablation study on illumination condition injector.}
\label{tb:illu}
\resizebox{\linewidth}{!}{ 
\begin{tabular}{l|cc|cc|cc|cc|cc}
\hline
Dataset  & \multicolumn{2}{c|}{LOL-Fog}  & \multicolumn{2}{c|}{LOL-Haze} & \multicolumn{2}{c|}{LOL-Rain} & \multicolumn{2}{c|}{LOL-Raindrop}  & \multicolumn{2}{c}{LOL-Snow} \\
Metrics  & PSNR & SSIM & PSNR & SSIM & PSNR & SSIM & PSNR & SSIM & PSNR & SSIM \\
\hline
w/o Illu. & 20.38 & 0.893 & 17.85 & 0.901  & 23.76 & 0.873 & 19.26 & 0.848 & 21.82 & 0.898 \\
Ours     & \textbf{29.34} & \textbf{0.956} & \textbf{27.74} & \textbf{0.981}  & \textbf{27.54}  & \textbf{0.903}& \textbf{26.03} & \textbf{0.922} & \textbf{27.50} & \textbf{0.938} \\
\hline
\end{tabular}
}
\end{table}

%% file: tex/table_tex/ablation/grid.tex
\begin{table}[t]
\centering
\caption{Ablation on grid step selection on LOL-V1}
\renewcommand{\arraystretch}{1.1}
\small
\label{tb:ablation_grid}
\begin{tabular}{@{} l|ccccc@{}}
\hline
Step & 4 & 8 & 16 & 32 & 64 \\
\hline
PSNR & 24.48 & 24.50 & \textbf{25.07} & 20.01 & 16.33 \\
SSIM & \textbf{0.92} & \textbf{0.92} & \textbf{0.92} & 0.89 & 0.83 \\
\hline
\end{tabular}
\end{table}

%% file: main/5_conclusion.tex
\section{Conclusion}

In this paper, we tackle the challenging task of restoring complex night scenes. To effectively perceive and model degraded patterns, we create comprehensive night scene datasets. It covers different weather degradations and illumination levels for end-to-end training.
To handle the complexity of night degradation, we introduce a new illumination-guided diffusion-based low-light image enhancement method. It uses an illumination map to enhance pixels dynamically with spatial variation. 
%
Extensive experiments on various benchmark datasets demonstrate the superior performance of our work.


